\pdfoutput=1
\documentclass[11pt,a4paper]{article}
\usepackage{amsmath}
\usepackage[hyperref]{acl2021}
\usepackage{times}
\usepackage{latexsym}

\usepackage{graphicx}
\usepackage{breqn}
\usepackage{amsfonts}
\usepackage{booktabs}
\usepackage{algorithm}
\usepackage[noend]{algpseudocode}

\usepackage{microtype}
\definecolor{ForestGreen}{RGB}{34,139,34}

\aclfinalcopy 

\newcommand{\joed}[1]{#1}

\newcommand{\joeed}[1]{}

\newcommand{\trevord}[1]{}

\title{Generating Diverse Descriptions from Semantic Graphs}

\author{Jiuzhou Han ~~~
  Daniel Beck ~~~
  Trevor Cohn \\
  School of Computing and Information Systems \\
  The University of Melbourne, Australia \\
  {\tt jiuzhouh@foxmail.com,} 
  {\tt \{d.beck,trevor.cohn\}@unimelb.edu.au}
}
\date{}

\begin{document}
\maketitle
\begin{abstract}

  Text generation from semantic graphs is traditionally performed with deterministic methods, which generate a unique description given an input graph. However, the generation problem admits a range of acceptable textual outputs, exhibiting lexical, syntactic and semantic variation. To address this disconnect, we present two main contributions. First, we propose a stochastic graph-to-text model, incorporating a latent variable in an encoder-decoder model, and its use in an ensemble. 
 Second, to assess the diversity of the generated sentences, we propose a new automatic evaluation metric which jointly evaluates output diversity and quality in a multi-reference setting.
We evaluate the models on WebNLG datasets in English and Russian, and show 
an ensemble of stochastic models produces diverse sets of generated sentences, while retaining similar quality to state-of-the-art models.



\end{abstract}

\section{Introduction}

Semantic graphs are an integral part of knowledge bases that integrate and store information in a structured and machine-accessible way \citep{DBLP:reference/fai/3}. 
They are usually limited to specific domains, describing concepts, entities and their relationships in the real world. Generating descriptions from semantic graphs is an important application of Natural Language Generation (NLG) and can be framed in a {\em graph-to-text} transduction approach. 

In recent years, approaches to graph-to-text generation can be broadly categorised into two groups. The first uses a sequence-to-sequence model \citep{DBLP:conf/acl/WangZQT18, DBLP:conf/acl/KonstasIYCZ17, DBLP:conf/emnlp/FerreiraLMK19}: the key step in this approach is to linearise the input graph to a sequence. Sequence-to-sequence models have been proved to be effective for tasks like question answering \citep{DBLP:conf/ijcai/YinJLSLL16}, text summarisation \citep{DBLP:conf/conll/NallapatiZSGX16}, and constituency parsing \citep{DBLP:conf/nips/VinyalsKKPSH15}. However, when dealing with graph inputs, this method does not take full advantage of the graph structure. Another approach is to handle the graph directly, using a graph-to-sequence model \citep{DBLP:journals/tacl/RibeiroZGG20, DBLP:conf/acl/CohnHB18, DBLP:conf/acl/ZhaoWC20}. This approach has been recently widely adopted as it shows better performance for generating text from graphs \citep{DBLP:journals/corr/abs-1804-00823}.

The models used in previous work are {\em deterministic}: given the same input graph, they will always generate the same text 
(assuming a deterministic decoding algorithm is used). 
However, it is widely known that many graphs admit multiple valid descriptions. This is evidenced by the presence of {\em multiple references} in datasets such as WebNLG \citep{DBLP:conf/acl/GardentSNP17, DBLP:conf/inlg/GardentSNP17} and it is a common phenomenon in other generation tasks such as machine translation and image captioning. In this work, we propose to use models that generate {\em sets of descriptions} instead of a single one. In particular, we develop {\em stochastic} models with latent variables that capture {\em diversity} aspects of semantic graph descriptions, such as lexical and syntactic variability. We also propose a novel evaluation methodology that combines quality and diversity into a single score, in order to address caveats of previously proposed diversity metrics. Our findings show that stochastic models perform favourably when generating sets of descriptions, without sacrificing the quality of state-of-the-art architectures.




\section{Related Work}

\paragraph{Graph-to-sequence Models}

Standard graph-to-sequence models have two main components: a graph encoder and a sequence decoder. The encoder learns the hidden representation of the input graph and the decoder generates text based on this representation. Different graph-to-sequence models vary mainly in the graph encoders.

\newcite{DBLP:conf/inlg/MarcheggianiP18} proposed an encoder based on Graph Convolutional Networks \citep[GCNs]{DBLP:conf/iclr/KipfW17}, which directly exploit the input structure. Similar to Convolutional Neural Networks \citep{lecun1998gradient}, GCN layers can be stacked, resulting in representations that take into account non-adjacent, long-distance neighbours. 
\newcite{DBLP:conf/acl/CohnHB18} used Gated Graph Neural Networks \citep{DBLP:journals/corr/LiTBZ15} by extending networks on graph architectures with gating mechanisms, similar to Gated Recurrent Units \citep[GRUs]{DBLP:conf/emnlp/ChoMGBBSB14}. \newcite{DBLP:conf/naacl/Koncel-Kedziorski19} proposed Graph Transformer Encoder by extending Transformers \citep{DBLP:conf/nips/VaswaniSPUJGKP17} to graph-structured inputs, based on the Graph Attention Network \citep[GAT]{DBLP:journals/corr/abs-1710-10903} architecture. This graph encoder generates node embeddings by attending over its neighbours through a self-attention strategy. \newcite{DBLP:journals/tacl/RibeiroZGG20} propose new models to encode an input graph with both global and local node contexts. To combine these two node representations together, they make a comparison between a cascaded architecture and a parallel architecture. 


\paragraph{Latent Variable Models}

Within neural networks, a standard approach for generative models with latent variables is the Variational Autoencoder (VAE) \citep{DBLP:journals/corr/KingmaW13}. 
The generative process is represented as: $p_{\theta}(x, z)=p_{\theta}(x \mid z) p_{\theta}(z)$, where $p_{\theta}(z)$ is the prior from which the latent variable is drawn, $p_{\theta}(x \mid z)$ is the likelihood of data point \textsl{x} conditioned on the latent variable $z$, typically calculated using a deep non-linear neural network, and $\theta$ denotes the model parameters.

\newcite{DBLP:conf/conll/BowmanVVDJB16} proposed a pioneering variational autoencoder 
for text generation to explicitly learn the global features using a continuous latent variable. They adapt the VAE to text data using an LSTM \citep{DBLP:journals/neco/HochreiterS97} for both the encoder and the decoder, using a Gaussian prior to build a sequence autoencoder. This architecture can be extended to {\em conditional} 
tasks (when there is an input guiding the generation). \newcite{DBLP:conf/emnlp/ZhangXSDZ16} proposed an end-to-end variational model for Neural Machine Translation (NMT), 
using a continuous latent variable to capture the semantics in source sentences and guide the translation process. 
\newcite{DBLP:conf/acl/CohnSA18} proposed a more expressive word-level 
machine translation model incorporating a chain of latent variables, modelling lexical and syntactic variation in parallel corpora.




\paragraph{Diversity in Neural Networks and Generation}


Variational latent variable models are commonly employed when there is a need for generating diverse outputs. This is achieved by sampling from the latent variable every time a new output is required. One can also use a standard deterministic model and sample from the decoder distributions instead but this tends to decrease the quality of the generated outputs. Here we review a few common techniques to address this issue.

Dropout \citep{DBLP:journals/jmlr/SrivastavaHKSS14} is a regularisation method used to prevent overfitting in neural networks. At training time, it masks random parameters in the network at every iteration. 
Dropout can also be employed in the testing phase, during generation. \joed{This idea was first proposed by \newcite{DBLP:conf/icml/GalG16} and it is also called Monte Carlo (MC) dropout.} Because MC dropout disables neurons randomly, the network will have different outputs every generation, which can make a deterministic model generate different outputs. 

Another technique to generate diverse outputs is {\em ensemble} learning. 
Typically, they are employed to prevent overfitting but they can also be used to generate diverse outputs. 
The idea is for each individual model in the ensemble to generate its own output. 
This approach can be very useful as each model tends to provide different optimal solutions in the network parameter space. This property has shown to benefit uncertainty estimation in deep learning \citep{DBLP:conf/nips/Lakshminarayanan17}. It can also be used both with deterministic and stochastic models, a property we exploit in our experiments.




\section{Stochastic Graph-to-Sequence Model}


\begin{figure*}[t]
    \centering
    \includegraphics[scale=0.6]{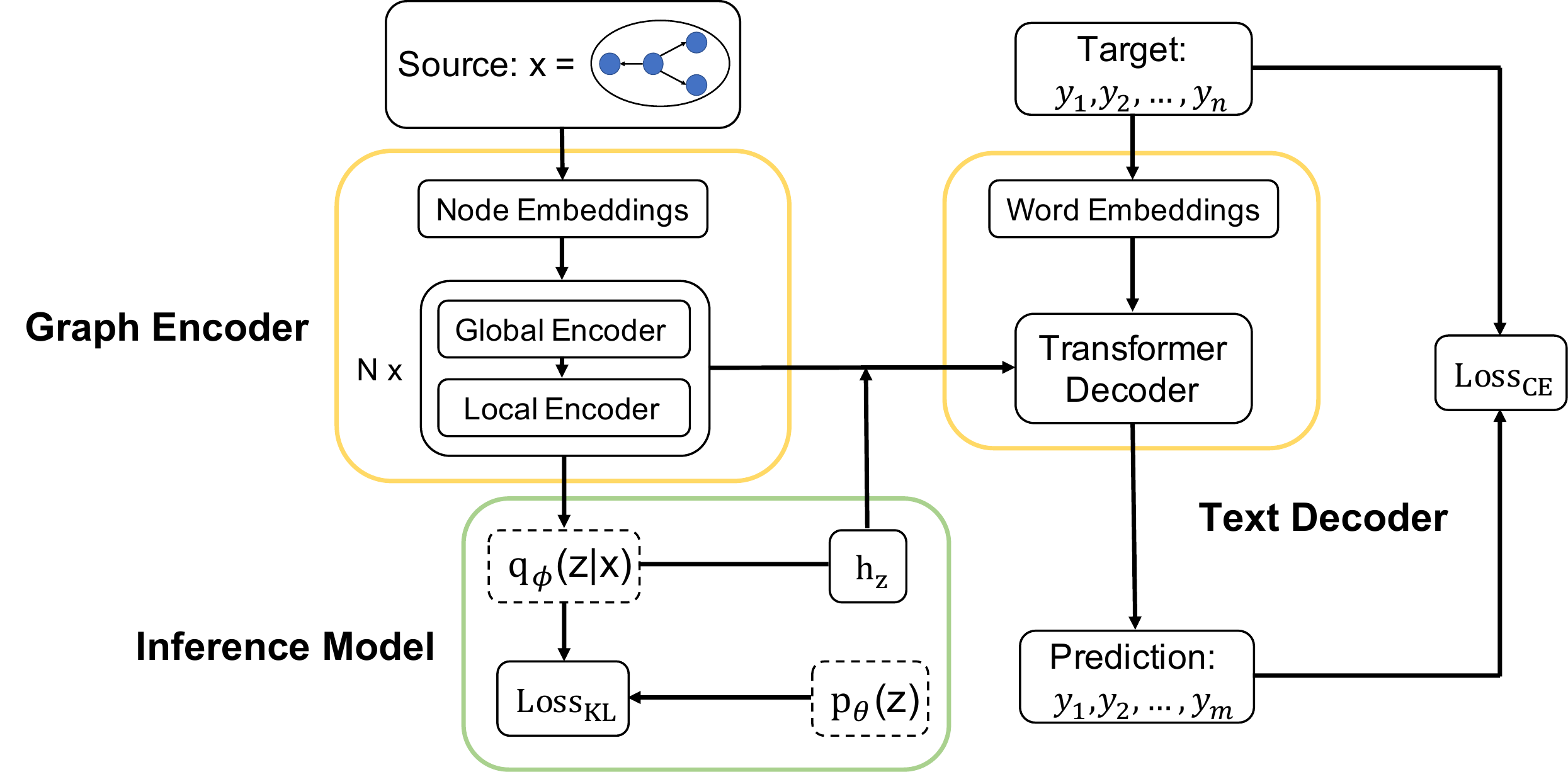}
    \caption{Proposed stochastic graph-to-sequence model architecture. \trevord{Update figure KLD $\rightarrow$ loss$_{KL}$; label $x$ = graph}}
    \label{fig:model}
\end{figure*}{}



In this section we introduce the proposed approach to generate diverse descriptions from semantic graphs. We start from the state-of-the-art model of \newcite{DBLP:journals/tacl/RibeiroZGG20}, which is a deterministic graph-to-sequence architecture. Then we incorporate a latent variable and a variational training procedure to this model, in order to turn the model stochastic. 
This latent variable aims at capturing linguistic variations in the descriptions and is responsible for increasing the diversity at generation time. The architecture is shown in Figure \ref{fig:model}.


\subsection{Graph Encoder and Text Decoder}

The encoder is similar to \newcite{DBLP:journals/tacl/RibeiroZGG20}, consisting of a {\em global} and a {\em local} subencoder. The global encoder considers a wide range of contexts but it ignores the graph topology by considering each node as if it were connected to all the other nodes in the graph. The local encoder learns the hidden representation of each node on the basis of its neighbour nodes, which exploits the graph structure effectively. Combining both global and local node aggregations, this encoder can learn better contextualised node embeddings. The global encoding strategy is mainly based on the Transformer architecture \citep{DBLP:conf/nips/VaswaniSPUJGKP17}, using a self-attention mechanism to calculate node representations of all nodes in the graph. The local encoding strategy adopts a modified version of Graph Attention Network \citep{DBLP:journals/corr/abs-1710-10903} by adding relational weights to calculate the local node representations.


The decoder is also based on a \emph{transformer} architecture. In our model, the input of the decoder is the contextualised node embeddings $h_{x}$ concatenated with the hidden state of the latent variable $h_{z}$, which can be represented as $[h_{x};h_{z}]$. Following \newcite{DBLP:journals/tacl/RibeiroZGG20}, we also use beam search with length penalty \citep{DBLP:journals/corr/WuSCLNMKCGMKSJL16} to encourage the model to generate longer sentences.


\subsection{Inference Model}

Here is where we introduce a latent Gaussian variable $z$, which together with the input graph $x$, guides the generation process. 
With this, the conditional probability of sentence $y$ given $x$ is
\begin{align*}
  p(y|x) 
  &= \int_z p(y|z,x) p(z|x) dz.
\end{align*}

The posterior inference in this model is intractable. Following previous work \citep{DBLP:conf/conll/BowmanVVDJB16, DBLP:journals/corr/KingmaW13}, we employ neural networks to fit the posterior distribution, to make the inference tractable. \joed{We regard} the posterior as a diagonal Gaussian $\mathcal{N}\left(\mu, \operatorname{diag}\left(\sigma^{2}\right)\right)$.
\trevord{Do we use the `standard' VAE? Not sure why the presentation isn't about our usage.}
The mean $\mu$ and variance $\sigma^{2}$ are parameterised with feed-forward neural networks (FFNNs), using the reparametrisation trick \citep{DBLP:conf/conll/BowmanVVDJB16, DBLP:journals/corr/KingmaW13} of the Gaussian variables. 
\trevord{From here on it's a bit of a tutorial, perhaps more depth than needed.}
It reparameterises the latent variable $z$ as a function of mean $\mu$ and variance $\sigma$: 
$$
z=\mu+\sigma \odot \epsilon \quad \epsilon \sim \mathcal{N}(0, \mathrm{I}),
$$
where $\epsilon$ is a standard Gaussian variable which plays the role of introducing noises, and $\odot$ denotes element-wise multiplication. The reparametrisation trick enables back-propagation in optimisation process with Stochastic Gradient Descent (SGD). Then we transform the latent variable $z$ into its hidden state $h_{z}$ through another FFNN.


The training objective encourages the model to keep its posterior distributions $q(z\mid\mathbf{x})$ close to a prior $p(z)$ that is a standard Gaussian $\mathcal{N}$($\mu = 0$, $\sigma = 1$). The loss function of the stochastic conditional model can be defined as  
\begin{multline*}
\mathcal{L}(\phi, \theta; \mathbf{x}, \mathbf{y})=-\mathbb{E}_{z \sim q_{\phi}\left(z \mid \mathbf{x}\right)}\left[\log p_{\theta}\left(\mathbf{y}\mid z, \mathbf{x}\right)\right] \\
+\mathrm{KL}\left(q_{\phi}\left(z \mid \mathbf{x}\right) \| p(z)\right).
\end{multline*}
The first term is the expected negative log-likelihood of data which is called reconstruction loss or cross-entropy loss. It forces the model to learn to reconstruct the data. The second term is the KL divergence which acts as a regulariser. By minimising the KL term, we want to make the approximate posterior stay close to the prior. We use SGD to optimise the loss function.




\subsection{Optimisation}

As shown above, the stochastic model objective comprises two terms reconstruction and KL regularisation. The KL divergence term will be non-zero and the cross-entropy term will be relatively small if the model encodes task-relevant information in the latent variable $z$. A difficulty of training is that the KL term tends to zero, causing the model to ignore $z$. This makes the model deterministic. This phenomenon is also known as the \emph{KL collapse} or \emph{KL vanishing problem} \citep{DBLP:conf/iclr/LucasTGN19}. We adopt the \emph{KL Threshold} method \cite{DBLP:journals/corr/abs-1812-04405}  to alleviate this issue. In this approach, we introduce a threshold $\zeta$ into the loss function to control the KL term. A large KL term means the latent variable learns much information. By setting a threshold, we can force the model to take at least a fixed KL regularisation cost. In our experiments, we set the threshold $\zeta$ as 10. The new loss function can be represented as
\begin{multline*}
\mathcal{L}(\phi, \theta; \mathbf{x}, \mathbf{y})=-\mathbb{E}_{z \sim q_{\phi}\left(z \mid \mathbf{x}\right)}\left[\log p_{\theta}\left(\mathbf{y}\mid z, \mathbf{x}\right)\right] \\ +\max \left(\mathrm{KL}\left(q_{\phi}\left(z \mid \mathbf{x}\right) \| p(z)\right), \zeta\right).
\end{multline*}





\section{Joint Evaluation of Diversity and Quality}
\label{proposed-metrics}

%



Addressing diversity in language generation is a recent topic that attracted attention in particular in image captioning. This led to the development of metrics that aim at measuring the diversity of a set of sentences, such as Self-BLEU \citep{DBLP:conf/sigir/ZhuLZGZWY18}. However, these metrics are based only on the generated output space, ignoring the references in the gold standard. This lead to spurious measurements, such as unconditional language models having excellent performance according to these metrics, even though they have no practical use as they ignore the input.

To address these caveats, we propose a new evaluation procedure that assesses diversity and quality {\em jointly}. Our key insight (and assumption) is based on using the reference set as a gold standard for both aspects. 
Given a graph, the set of references acts as the ``end goal'', containing high-quality descriptions with sufficient levels of diversity.\footnote{We discuss limitations of this assumption in Section \ref{sec:conc}.} We call this procedure {\bf Multi-Score (MS)}. 

\begin{figure}[t]
    \centering
    \includegraphics[scale=0.6]{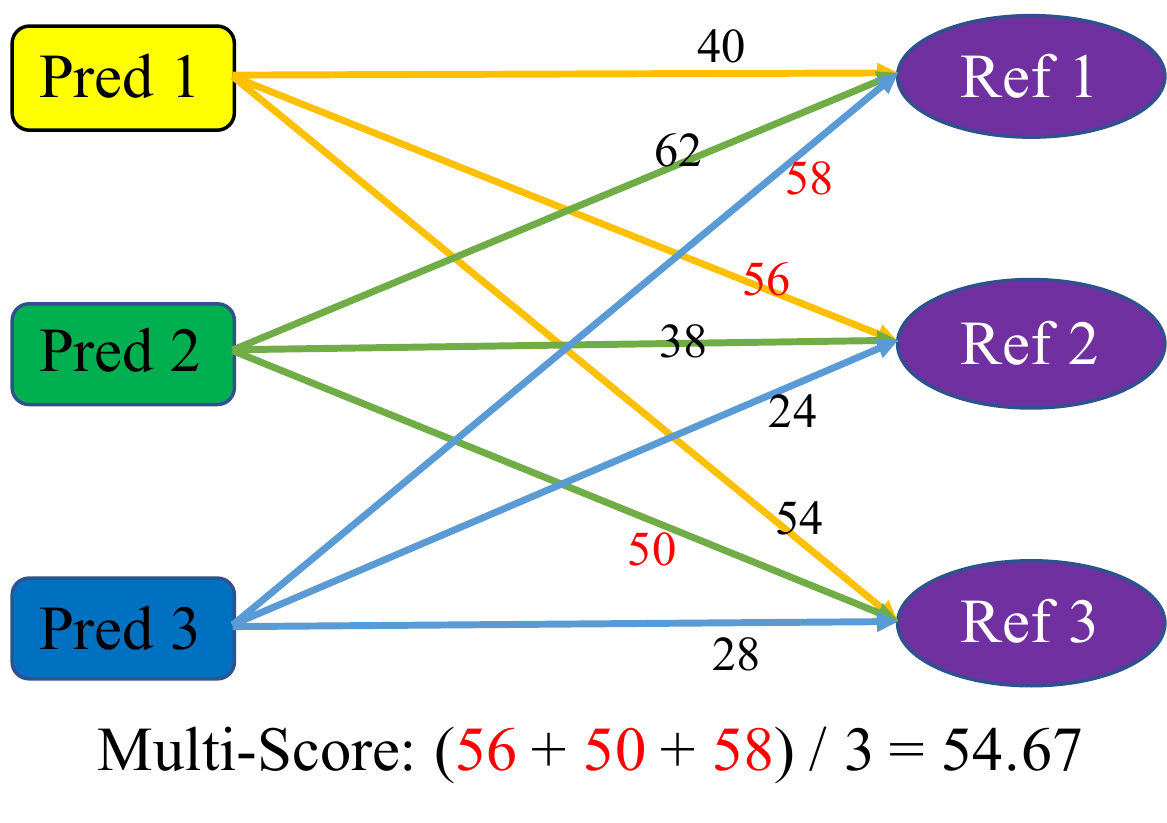}
    \caption{An example of calculating Multi-Score. The three ``Pred'' nodes on the left side represent three predicted descriptions while the three ``Ref'' nodes on the right side represent three references. The weight of each edge corresponds to the sentence-level quality score of this prediction-reference pair. The scores highlighted in red are the ones corresponding to the maximal matching, which are then used to calculate the MS metric. Other scores are ignored.}
    \label{fig:multiscore}
\end{figure}

The idea behind Multi-Score is shown pictorially in Figure \ref{fig:multiscore}. In this example, we have a single instance with three references and three predicted descriptions generated by a model. Given a sentence-level quality metric 
we can calculate it among {\em all possible pairs} between each prediction and reference, obtaining a weighted bipartite graph. We then solve the respective {\em maximum matching problem} for this bipartite graph and take the average weight of the edges corresponding to the optimal matching. We show the full procedure to calculate Multi-Score in Algorithm \ref{alg:ms}.


\begin{algorithm}
  \begin{algorithmic} 
    \Function{Multi-Score}{${\bf o}$: outputs, ${\bf r}$: references, $\mathcal{M}$: sentence-level metric}
    \State ${\bf G} \leftarrow {\bf 0}$ \Comment{initialise graph}
    \For{$i \leftarrow 0 \text{ to len}({\bf o})$} \Comment{fill graph}
    \For{$j \leftarrow 0 \text{ to len}({\bf r})$}
    \State ${\bf G}(i,j) \leftarrow \mathcal{M}({\bf o}[i],{\bf r}[j])$
    \EndFor
    \EndFor
    \State ${\bf match} \leftarrow $ {\sc MaxMatch}(${\bf G}$) \Comment{stores edges}
    \State score $\leftarrow 0$
    \For{edge $\in {\bf match}$}
    \State score $\leftarrow$ score + edge.weight
    \EndFor
    \State \Return score / len(${\bf match}$)
    \State \Comment{returns average weight}
    \EndFunction
\end{algorithmic}
\caption{Multi-Score procedure}
\label{alg:ms}
\end{algorithm}

For the example in Figure \ref{fig:multiscore}, the optimal matching (shown in red) matches prediction 1 with output 2, prediction 2 with output 3 and prediction 3 with output 1. From this, the resulting Multi-Score is:
$({56}+{50}+{58})/3 = {\bf 54.67}$. The matching problem {\sc MaxMatch} can be solved using the Hungarian Algorithm \citep{DBLP:books/daglib/p/Kuhn10} in $O(n^3)$ time, where $n$ is the number of nodes in the bipartite graph. This makes the procedure efficient for reference set sizes found in standard datasets.

As a metric, Multi-Score has a number of desirable properties:
\begin{itemize}
\item As long as the sentence-level metric has an upper bound (which is the case of most standard automatic evaluation metrics), if the set of predictions is exactly equal to the references, then MS will give the maximum score.
\item If the outputs are diverse but unrelated to the references (as in an unconditional LM), MS will penalise the output because the underlying quality values will be low.
\item If the outputs are high-quality but not diverse (typical of an n-best list in a deterministic model), MS will penalise the output due to the assignment constraint. One of the outputs will have a high-quality value but the others will have a low-quality value because they will be forced to match other references.
\item Finally, MS can be used with any sentence-level quality metric, making it easily adaptable to any developments in better quality metrics, as well as other generation tasks.
\end{itemize}

\section{Experimental Settings}

\subsection{Dataset}

We evaluate the models using datasets from the WebNLG shared tasks \citep{DBLP:conf/acl/GardentSNP17, DBLP:conf/inlg/GardentSNP17}. The data is composed of data-text pairs where the data is a set of RDF triples extracted from DBpedia and the text is the verbalisation of these triples. 
For each graph, there may be multiple descriptions. In our experiments, we assume a reference set of size 3 for each input, as most graphs in both datasets have three reference descriptions.

\paragraph{English WebNLG 2017}

This dataset contains 18102 training, 872 development and 971 test data-text pairs. Entities are classified into 15 distinct categories (Astronaut, University, Monument, Building, ComicsCharacter, Food, Airport, SportsTeam, WrittenWork, Athlete, Artist, City, MeanOfTransportation, CelestialBody, Politician). 

\paragraph{Russian WebNLG 2020}
The Russian dataset comprises 16571 training, 790 development and 1102 test data-text pairs. This dataset has 9 distinct categories (Airport, Astronaut, Building, CelestialBody, ComicsCharacter, Food, Monument, SportsTeam, and University).

\subsection{Preprocessing}

\paragraph{Levi Graph Transformation}

To decrease the number of parameters and avoid parameter explosion, we follow previous work and use a Levi Graph Transformation \citep{DBLP:journals/tacl/RibeiroZGG20, DBLP:conf/acl/CohnHB18}. 
This transformation creates new relation nodes from relational edges between entities, which explicitly represents the relations between an original node and its neighbour edges. 

\paragraph{Byte Pair Encoding}

Following previous work \citep{DBLP:journals/tacl/RibeiroZGG20}, we employ Byte Pair Encoding (BPE) to split entity words into frequent characters or character sequences which are subword units. 
\trevord{And each node in the graph is a phrase (not a word)?}
After the BPE operations, some nodes in the graph are 
split to subwords. Likewise, we also split the target descriptions using BPE. 

\subsection{Models}

All models are able to generate sets of descriptions: we generate three sentences per graph as this matches the number of available references. For the proposed stochastic models, we generate each sentence by sampling a new value for the latent variable. For the deterministic models, we use different decoding strategies to generate these sets. 


\paragraph{Top-3 Beam Search}

Beam Search is the standard algorithm to obtain a sentence from deterministic models, by selecting the output with (approximate) highest probability. In Top-3 Beam Search, we choose the top-3 generated sentences from the final candidate list. 

\paragraph{Total Random Sampling}

Random Sampling \citep{DBLP:conf/acl/IppolitoKSKC19} generates a sentence from left to right 
sampling the next token from all possible candidates until the end-of-sequence symbol is generated. Because each token is {\em sampled} from the distribution over next tokens given the previous ones, this method generates different outputs each time it generates a new description.
\trevord{I take it that each word is generated from the distribution over next words under the predictive model? That is, it's not uniform.}

\paragraph{Top-3 Random Sampling}

In this approach, we still use Random Sampling but modify it slightly while generating the next token. Instead of sampling the next token from all possible candidates, the model samples the next token from the top-3 most likely candidates \citep{DBLP:conf/acl/IppolitoKSKC19}. 
\trevord{I take it that each word is generated uniformly?}

\joed{\paragraph{MC Dropout} We employ MC dropout to the deterministic model and keep the dropout rate in the testing phase and training phase the same. It disables neurons randomly at decoding time, resulting in different outputs at each generation.} 

\paragraph{Ensemble}

Finally, we create an ensemble of three independently-trained deterministic models, whereby we select the most likely sentence from each model using Beam Search. These sentences then form the output set from the ensemble. Since this is a general strategy, {\em we also apply it to the stochastic model} as another point of comparison in our experiments.

\joeed{Since we talk about ensemble model here, do we still need to include the test dropout in this part? I have added.}




\section{Results}


We assess each model on the test set of English and Russian datasets respectively and report the quality and diversity results. 
The quality evaluation scores (BLEU: \citet{DBLP:conf/acl/PapineniRWZ02}, CHRF++: \citet{DBLP:conf/wmt/Popovic17}) are calculated based on the average score of the three outputs. We report the original BLEU and CHRF++ score to show the quality of the generated sentences from each model. The diversity evaluation scores (Self-BLEU, Multi-Score) are computed using the three outputs. As we describe in Section \ref{proposed-metrics}, our proposed diversity evaluation metrics require a sentence-level quality evaluation metric to compute the score of two sentences. We adopt sentence-level BLEU and CHRF++ and refer to their corresponding Multi-Score versions as MS-BLEU and MS-CHRF.

\begin{table*}[t]
  \centering
  \begin{tabular}{lcccc}
    \toprule
    & \multicolumn{2}{c}{\bf English} & \multicolumn{2}{c}{\bf Russian}\\
    & \textbf{BLEU↑} & \textbf{CHRF++↑} & \textbf{BLEU↑} & \textbf{CHRF++↑} \\
    \midrule
    \multicolumn{5}{l}{\bf Deterministic Models} \\
    Top-3 beam search               & 62.69         & 74.48   & 52.50         & 64.76       \\
    Total random sampling           & 49.01         & 66.35    & 40.62         & 57.06       \\
    Top-3 random sampling           & 56.62         & 71.16    & 46.91         & 61.45       \\
    MC dropout                      & 59.10         & 71.57    & 47.97         & 61.41       \\
    Ensemble                        & \bf 63.31         & \bf 74.52    & \bf 53.60         & \bf 65.30       \\
    \midrule
    \multicolumn{5}{l}{\bf Stochastic Models} \\
    Single model            & 62.81         & 74.12           & 52.45         & 64.43           \\
    Ensemble          & 62.88         & 74.25           & 52.60         & 64.38           \\ 
    \midrule
    \multicolumn{5}{l}{\bf Previous Work} \\
    Melbourne \citep{DBLP:conf/inlg/GardentSNP17}                      & 54.52         & 70.72  & - & -         \\
    Adapt \citep{DBLP:conf/inlg/GardentSNP17}                           & 60.59         & 76.01 & - & -          \\
    CGE-LW \citep{DBLP:journals/tacl/RibeiroZGG20}                         & 63.69         & 76.66  & - & -         \\
    \bottomrule
  \end{tabular}
  \caption{Quality evaluation results on the test sets of both English and Russian datasets. Note that models without declaring decoding strategy use Beam Search. For reference, we also report results from previous work in the English dataset. 
  Boldface shows the best result for a column, and arrows indicate the direction of improvement, i.e., $\uparrow$: higher is better.
  } 
  \label{tab:quality}
\end{table*}

\begin{table*}[t!]
  \centering
\begin{tabular}{lccccccc}
  \toprule
  & \multicolumn{3}{c}{\bf English} & \multicolumn{3}{c}{\bf Russian}\\
  & \textbf{Self-B$\downarrow$}      & \textbf{MS-B$\uparrow$} & \textbf{MS-C$\uparrow$} & \textbf{Self-B$\downarrow$}      & \textbf{MS-B$\uparrow$} & \textbf{MS-C$\uparrow$} \\
  \midrule
  \multicolumn{7}{l}{\bf Deterministic Models} \\
  Top-3 beam search               & 86.72          & 46.65               & 71.45      & 76.50               & 38.23               & 61.58         \\
  Total random sampling           & \textbf{56.48} & 40.47               & 67.00      & \textbf{52.30}      & 31.37               & 56.30         \\
  Top-3 random sampling           & 64.66          & 45.15               & 70.40      & 60.31              & 35.61               & 59.95          \\
  MC dropout                     & 68.70           & 46.90               & 70.87     & 61.59              & 36.14               & 59.37           \\
  Ensemble                        & 81.31          & 47.32               & 71.52      & 75.70              & 38.50                & 61.71         \\
  \midrule
  \multicolumn{5}{l}{\bf Stochastic Models} \\
  Single model & 97.30          & 43.25               & 69.45  & 97.62              & 33.53               & 58.40             \\
  Ensemble  & 77.85          & \textbf{47.61}      & \textbf{71.95}  & 73.50              & \textbf{38.86}               & \textbf{61.95}     \\
  \bottomrule
\end{tabular}
\caption{Diversity evaluation results on the test sets of both English and Russian datasets. {\bf Self-B} refers to Self-BLEU while {\bf MS-B} and {\bf MS-C} refer to the proposed Multi-Score metric using sentence-level BLEU and CHRF++ as the underlying quality metric. Note that models without declaring decoding strategy use beam search decoding.}
\label{tab:diversity}
\end{table*}

Table \ref{tab:quality} shows the quality results on both English and Russian datasets. 
\joed{As expected, the two random sampling methods do not show good quality performance. 
For English data, our stochastic models perform on par with previous work and have comparable quality with deterministic models. The trends for English and Russian data are similar but Russian has lower scores in general. 
}

The diversity scores of these two datasets are shown in Table \ref{tab:diversity}. 
\joed{\emph{Total random sampling}  has the lowest Self-BLEU on two datasets, as expected, but it also has the worst quality. On the other hand, with our new metrics, the stochastic ensemble model gives the best results on both English and Russian datasets, showing high diversity without compromising quality.}


\subsection{Error Analysis}



To further assess the quality of the generated sentences from each model, we perform a manual error analysis in a subset of the English test data. We randomly selected five input graphs, generating 15 sentences for each model (as we generate 3 sentences for each graph). Given we analysed five models, this gives a total of 75 sentences for our analysis. We observed three common mistakes from the outputs:

\begin{itemize}
    \item \textbf{Syntax/Spelling Mistake:} There are grammar mistakes or spelling mistakes. 
    \item \textbf{Lack of Information:} The information in the graph is not fully realised in the description.
    \item \textbf{Information Redundancy:} Some information in the sentence is repeated.
\end{itemize}

\begin{table*}[t!]
\centering
\scalebox{1.0}{%
\begin{tabular}{lcccc}
\toprule
  \textbf{}       \textbf{Models}          & \textbf{\begin{tabular}[c]{@{}c@{}}Syntax/Spelling\\ Mistake\end{tabular}} & \textbf{\begin{tabular}[c]{@{}c@{}}Lack of\\ Information\end{tabular}} & \textbf{\begin{tabular}[c]{@{}c@{}}Information\\ Redundancy\end{tabular}} & \textbf{Average} \\
  \midrule
  \multicolumn{5}{l}{\bf Deterministic Models} \\
Total random sampling & 0.54                                                                         & 0.18                                                                                                                                            & 0.20                                                                         & 0.33             \\
Top-3 random sampling & 0.18                                                                          & 0.14                                                                                                                                         & 0.49                                                                         & 0.22             \\
MC dropout          & 0.18                                                                          & 0.32                                                                                                                                          & 0.20                                                                         & 0.22             \\
Top-3 beam search     & 0.07                                                                          & 0.14                                                                                                                                            & 0.00                                                                         & 0.09              \\
Ensemble     & 0.00                                                                         & 0.09                                                                                                                                            & 0.03                                                                         & 0.06              \\
    \multicolumn{5}{l}{\bf Stochastic Models} \\
Ensemble                        & 0.03                                                                          & 0.13                                                                                                             & 0.08                                                                     & 0.08              \\ \bottomrule
\end{tabular}}
\caption{Error analysis results, showing the rates of mistakes for each model.}
\label{tab:stat}
\end{table*}

\begin{table*}[t]
  \centering
    \begin{small}
\begin{tabular}{p{0.9\textwidth}}
\hline 
\begin{minipage}{0.9\textwidth}
    \centering 
    \includegraphics[scale=0.6]{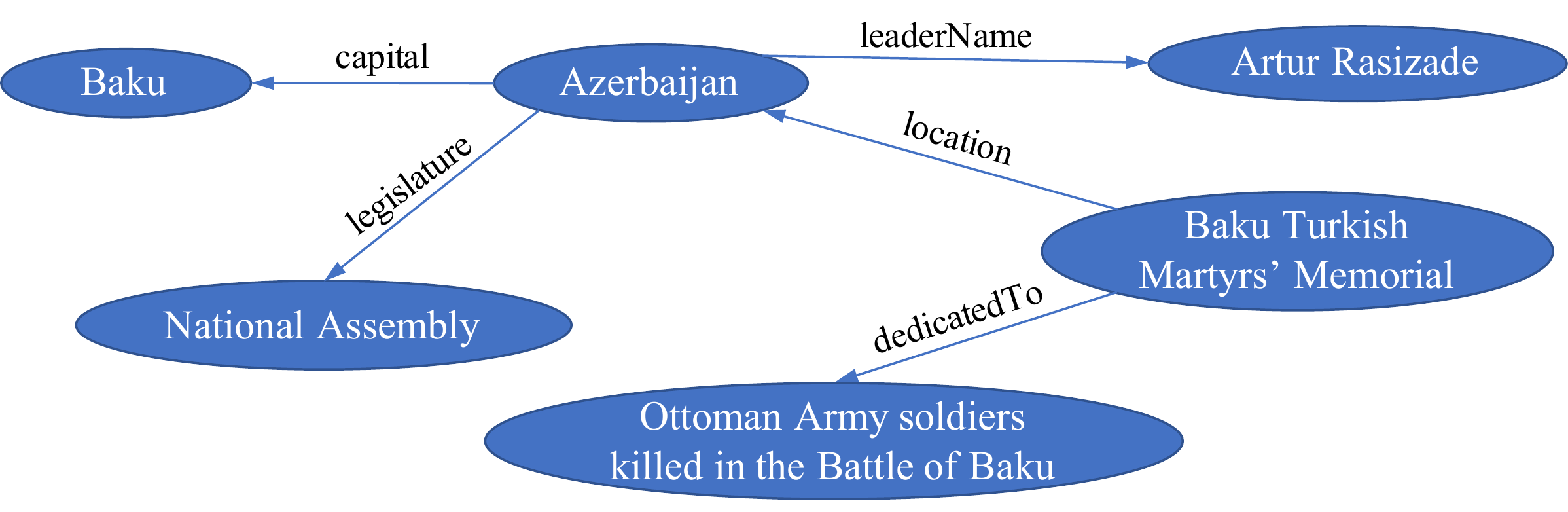}
\end{minipage} \\
  \hline

\textsl{DM (MC dropout) 1:} The Baku Turkish Martyrs’ Memorial, which is dedicated to the Ottoman Army soldiers killed in the battle of Baku, is found in Azerbaijan. The capital of Azerbaijan is Baku and the leader is Artur Rasizade. \textcolor{ForestGreen}{(missing: legislature information)} \\
\textsl{DM (MC dropout) 2:} The Baku Turkish Martyrs’ Memorial, which is dedicated to the Ottoman Army soldiers killed in the battle of Baku, \textcolor{orange}{is dedicated to the Ottoman Army soldiers killed in the country} \textcolor{red}{is led by Artur Rasizade}. \textcolor{ForestGreen}{(missing: legislature information)} \\
\textsl{DM (MC dropout) 3:} The Baku Turkish Martyrs’ Memorial is dedicated to the Ottoman Army soldiers killed in the battle of Baku. \textcolor{orange}{It is dedicated to the Ottoman Army soldiers killed in the battle of Baku,} the leader of Azerbaijan is Artur Rasizade. \textcolor{ForestGreen}{(missing: legislature information)} \\
\hline
\textsl{SM (Ensemble) 1:} The Baku Turkish Martyrs’ Memorial is dedicated to the Ottoman Army soldiers killed in the battle of Baku. It \textbf{is located in} Azerbaijan \textbf{whose capital is} Baku and its leader is Artur Rasizade. \textbf{The legislature is} the National Assembly. \\
\textsl{SM (Ensemble) 2:} Baku \textbf{is the capital of} Azerbaijan \textbf{where the legislature is} the National Assembly and the leader is Artur Rasizade. The country \textbf{is the location of} the Baku Turkish Martyrs Memorial \textbf{which is dedicated to} the Ottoman Army soldiers killed in the battle of Baku. \\
\textsl{SM (Ensemble) 3:} The Baku Turkish Martyrs’ Memorial \textbf{is dedicated to} the Ottoman Army soldiers killed in the battle of Baku. It is located in Azerbaijan whose capital is Baku and its leader is Artur Rasizade\textbf{, and its legislature is} the National Assembly. \\
\hline
\end{tabular}
\end{small}
\caption{A WebNLG input graph and the outputs from a Deterministic Model (MC dropout) and a Stochastic Model (Ensemble). We use different coloured texts to indicate four different mistakes: \textcolor{red}{red} colour represents Syntax/Spelling Mistake, \textcolor{ForestGreen}{green} colour represents Lack of Information, \textcolor{blue}{blue} colour represents Inaccurate Expression, \textcolor{orange}{orange} colour represents Information Redundancy. \textbf{Bold} texts show syntactic variations.}
\label{tablecase3}
\end{table*}

We calculate the rates of each model making different types of mistakes and report the results in Table \ref{tab:stat}. The results show that total random sampling makes the most mistakes among all models and most of them are syntax or spelling mistakes. \emph{Top-3 random sampling} and \emph{MC dropout} make the same percentage of total mistakes. The former makes almost half of the total information redundancy mistakes while the latter makes the most lack of information mistakes. Top-3 beam search makes fewer mistakes than the other three models and it does not make information redundancy mistakes in our evaluated test cases.

\joed{As for ensemble-based models, both deterministic and stochastic ensembles make the fewest total mistakes among all models. This is in line with the results obtained from automatic quality metrics. 
In particular, the deterministic ensemble does not make any syntax or spelling mistakes in the evaluated test cases. The stochastic ensemble also shows good performance with regard to the quality of the generated sentences, which has a low error rate for all types of mistakes.}

In general, the diverse outputs generated by our proposed model tend to have comparable quality to the outputs from the best baseline model. However, lack of information still remains a challenge for some instances in this setting. Addressing this problem is an avenue that we leave for future work.

\subsection{Case Study}
Table \ref{tablecase3} shows an instance of a semantic graph from which we collect three outputs from a deterministic model (MC dropout) and a stochastic model (Ensemble). The outputs from MC dropout contain three types of mistakes and have low diversity. While there is no mistake in the outputs of the stochastic model, and the boldface illustrates syntactic variation.

\section{Conclusion and Future Work}
\label{sec:conc}


In this work, we first propose stochastic graph-to-text models to generate diverse sentences from semantic graphs. This was implemented through latent variable models that aim to capture linguistic variation and ensembling techniques. 
Furthermore, to solve the limitation of the existing diversity evaluation metrics, we also propose Multi-Score, a new automatic evaluation metric assessing diversity and quality jointly. It provides a general and effective way to assess the diversity of generated sentences for any text generation task. We perform experiments on English and Russian datasets and results demonstrate the generated sentences from the stochastic ensemble have both high diversity and high quality.

Since Multi-Score is based on using the reference set as the gold standard, it has a limitation that the variety of the reference sentences can largely influence the metric. Datasets containing reference sentences with higher quality and diversity will likely generate a more accurate Multi-Score for the predicted sentences. In other words, Multi-Score evaluates diversity {\em implicitly} through the references, as opposed to {\em explicit} judgements of diversity. However, explicit human evaluation requires a formal definition of diversity which is difficult to establish (as compared to quality judgements, for instance). Nevertheless, addressing this challenge could provide a pathway to reduce the need for multiple references in evaluating diversity. 

To the best of our knowledge this is the first work that incorporates a latent variable within a graph-to-sequence model. This in turn leads to many promising research avenues to explore in future work. Our analysis showed that the latent variable mostly helps in syntactic variation but less in other aspects such as semantics. Analysing the behaviour of the latent variable when modelling linguistic information is an important avenue that will enhance the understanding of stochastic models.





\bibliographystyle{acl_natbib}
\bibliography{acl2021}


\end{document}